\documentclass{SCIS2023}
\usepackage{multirow}
\usepackage{subcaption}
\usepackage{graphicx}
\usepackage{booktabs}
\usepackage[export]{adjustbox}
\usepackage{caption}
\usepackage{diagbox} 
\usepackage{slashbox}
\usepackage{makecell}
\usepackage{array} 
\usepackage{subcaption}
\usepackage{xspace}
\usepackage{cleveref}
\usepackage{hyperref}
\usepackage{hypcap}

\newcommand{\methodname}{CPF\xspace}
\newcommand{\Qmatrix}{Q-matrix\xspace}
\newcommand{\Pmatrix}{P-matrix\xspace}

\begin{document}
\ArticleType{RESEARCH PAPER}
\Year{2022}
\Month{}
\Vol{}
\No{}
\DOI{}
\ArtNo{}
\ReceiveDate{}
\ReviseDate{}
\AcceptDate{}
\OnlineDate{}


\title{Personalized Forgetting Mechanism with Concept-Driven Knowledge Tracing}{Title keyword 5 for citation Title for citation Title for citation}

\author[1]{Shanshan WANG}{}
\author[2]{Ying HU}{}
\author[3]{Xun YANG}{}
\author[4]{Zhongzhou ZHANG}{}
\author[5]{\\Keyang WANG}{}
\author[6]{Xingyi ZHANG}{}

\AuthorMark{}

\AuthorCitation{}


\address{\textsuperscript{1,2}Information Materials and Intelligent Sensing Laboratory of Anhui Province, Anhui University, Anhui {\rm 230601}, China}
\address[3]{School of Data Science, University of Science and Technology of China, Anhui {\rm 230026}, China}
\address[4]{ College of Computer Science, Sichuan University, Chengdu {\rm 610065}, China}
\address[5]{zhejiang Dahua Technology Co., Ltd. Zhejiang {\rm 310053}, China}
\address[6]{School of Computer Science and Technology, Anhui University, Anhui {\rm 230601}, China}

\abstract{
Knowledge Tracing (KT) aims to trace changes in students’ knowledge states throughout their entire learning process by analyzing their historical learning data and predicting their future learning performance. Existing forgetting curve theory based knowledge tracing models only consider the general forgetting caused by time intervals, ignoring the individualization of students and the causal relationship of the forgetting process. To address these problems, we propose a \textbf{C}oncept-driven \textbf{P}ersonalized \textbf{F}orgetting knowledge tracing model (CPF) which integrates hierarchical relationships between knowledge concepts and incorporates students’ personalized cognitive abilities. First, we integrate the students' personalized capabilities into both the learning and forgetting processes to explicitly distinguish students’ individual learning gains and forgetting rates according to their cognitive abilities. Second, we take into account the hierarchical relationships between knowledge points and design a precursor-successor knowledge concept matrix to simulate the causal relationship in the forgetting process, while also integrating the potential impact of forgetting prior knowledge points on subsequent ones. The proposed personalized forgetting mechanism can not only be applied to the learning of specifc knowledge concepts but also the life-long learning process. Extensive experimental results on three public datasets show that our CPF outperforms current forgetting curve theory based methods in predicting student performance, demonstrating CPF can better simulate changes in students’ knowledge status through the personalized forgetting mechanism.

}


\maketitle 
\section{Introduction}
With the rapid development of online education, it has become increasingly important to accurately assess students’ knowledge status. 
Knowledge Tracing (KT) plays a vital role in online learning, which is the task of modelling student knowledge over time so that we can accurately predict the student's performance on future interactions.
Knowledge tracing~\cite{anderson1990cognitive,villano1992probabilistic} models not only help educators better understand students' academic abilities ~\cite{nakagawa2019graph} but also provide targeted educational resources and personalized learning strategies. By analyzing students' historical learning data, \textit{i.e.}, learning behaviors and performance~\cite{Ma2019}, they can accurately predict their performance and knowledge level on future learning tasks~\cite{Bai2023ArgusDroid}. Considering the high cost and importance of data collection in online education, the use of knowledge tracing models is expected to grow in prevalence and importance within the field of education.

\begin{figure}[h]
    \centering
    \includegraphics[width=1.0\linewidth]{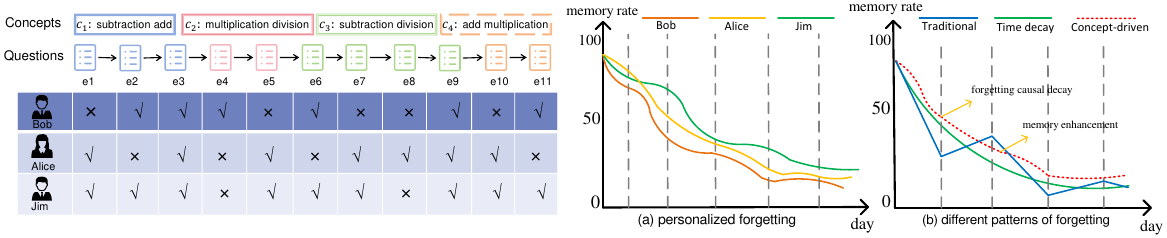}
    \caption{Examples of three learners answer a series of exercises on an online learning system. (a) indicates that the forgetting rate of various students is different. (b) represents different forgetting patterns: the traditional model considers correct answers as mastering the knowledge point, while incorrect answers indicate no impression of the knowledge point; the time decay model believes that memory of knowledge points weakens over time; and the causal forgetting model integrates the intrinsic relationship between time and knowledge points, balancing long-term and short-term memory.}
    \label{fig:model_architecture}
     \label{fig:forget}
\end{figure}

Existing knowledge tracing (KT) models~\cite {Habimana2020SentimentAnalysis, Su2023ModularNN} have achieved tremendous success in predicting students' performance. For example, Deep Knowledge Tracing (DKT)~\cite{10.5555/2969239.2969296, Moerland2023ModelBasedRL} utilizes Recurrent Neural Networks (RNN) and Long Short-Term Memory (LSTM) networks~\cite{Huang2021, Xie2020,Su2023ModularNN} to predict correct answers.
During the practice process, once a student answers incorrectly, DKT argues that the student's corresponding knowledge status will decline. 
Dynamic Key-Value Memory Network (DKVMN)~\cite{10.1145/3038912.3052580} introduces a dynamic key-value memory matrix based on DKT to capture changes in students' knowledge state more efficiently. 
Context-aware Knowledge Tracing (AKT)~\cite{Ghosh2020ContextAwareAK} is an attention-based model that connects a learner's future responses to assess questions with their past responses. It calculates attention weights using exponential decay and context-aware relative distance metrics, alongside question similarity.
Learning Process-consistent Knowledge Tracing (LPKT)~\cite{10.1145/3447548.3467237} monitors students' knowledge status by directly modeling the student's learning process, calculates learning gain through the difference between learning units, and simulates the forgetting process through time intervals~\cite{10.1145/3379507}.
Considering that progress rates are student-specific, extending LPKT to LPKT-S~\cite{9950313} differentiates individual progress rates for each student. 
Although the aforementioned methods have successfully modeled the forgetting process, there is still a significant limitation that they fail to consider the personalized cognitive processes of students and the causal relationships of the forgetting process. As a result, this limitation weakens the interpretability ~\cite{Xie2017TopicEnhanced} of these methods, making it difficult for educators to gain deep insights and limits a comprehensive understanding of students’ learning and forgetting processes. Therefore, there is a need to meticulously and comprehensively consider the cognitive structure and knowledge structure of students in the KT task. As shown in Fig.~\ref{fig:forget}, the different cognitive structures of students will not only lead to different learning outcomes, but also different forgetting rates. Generally, students with stronger learning abilities can acquire knowledge more eﬀectively in the learning process and have lower forgetting rates in the forgetting process. Conversely, students with weaker learning abilities tend to have poorer grasp of knowledge and also exhibit higher rates of forgetting. Additionally, the different knowledge structures can affect the students' mastery of the knowledge because there is a hierarchical relationship between knowledge concepts. Specifically, if one knowledge point is forgotten by the student, the related knowledge points may also be forgotten to some extent.

To better understand the causal relationship between personalized learning and the forgetting process of students, in this paper, we propose a Concept-driven Personalized Forgetting knowledge tracing (\methodname) model. Firstly, to accommodate the individual diﬀerences among students, we incorporate students’ prior abilities into the modeling of both the learning and forgetting processes. Guided by cognitive theory~\cite{ECN2019} and learning efficacy theory~\cite{Bai2021UsingST}, we measure students’ abilities~\cite{8909656} through answer time, answer accuracy and question difficulty in this study. Secondly, to explore the causal relationship between knowledge concepts in the forgetting process, we design a novel causal forgetting mechanism that explicitly quantify the hierarchy of different knowledge concepts. Specifically, inspired by educational theories~\cite{Geng2017ModelBasedDiagnosis,Zhang2023}, we construct a precursor-successor matrix (\Pmatrix) to capture the directed relations between diﬀerent knowledge concepts and calculate causal forgetting weights by searching for the closest prerequisite knowledge concepts to the current exercise according to this \Pmatrix. Subsequently, to simulate the forgetting-review mechanism presented by students when facing the exercise with related knowledge concepts, we compute the similarity of adjacent knowledge states and adopt this similarity to update the forgetting gate, thereby modeling the forgetting process more accurately~\cite{Ji2022HeterogeneousMemory,arxiv:2307.00811} in the real-world scenarios. We conduct on three public datasets used in KT task. The state-of-the-art performance demonstrates the effectiveness of our \methodname.

The main contributions of our paper are as follows:
\begin{itemize}
    \item We incorporate students’ personalized cognitive abilities into both the learning and forgetting processes in the knowledge tracing, which can explicitly distinguish students’ individual learning gains and forgetting rates according to their cognitive abilities.
    \item We design a precursor-successor knowledge concept matrix and introduce a forgetting-review mechanism, aiming to comprehensively capture the causal relationships between knowledge concepts during the forgetting process of knowledge tracing.
    \item By integrating the personalized cognitive processes and the causal relationships of concepts in the forgetting process, a novel Concept-driven Personalized Forgetting knowledge tracing (\methodname) model is materialized in this paper. Extensive experimental results on three public datasets demonstrate the effectiveness of our CFP.
\end{itemize}

\section{ Related Works}
\subsection{Knowledge Tracing}
Knowledge tracing (KT)~\cite{Zhao2024} is an important task in online learning systems, aiming to dynamically track the learner's knowledge status. Existing methods can be divided into two categories: (1) Traditional methods, such as Bayesian Knowledge Tracing (BKT)~\cite{corbett1994knowledge,Gao2021,Wang2013Restricted}, which use binary variables to describe students' knowledge status. The Item Response Theory model (IRT)~\cite{Lord1980,doi:https://doi.org/10.1002/9781118625392.wbecp357} uses student abilities and problem characteristics to analyze student performance. Traditional KT models ignore the forgetting process. (2) Based on deep learning methods, Deep Knowledge Tracing (DKT) introduces recurrent neural networks such as RNN and LSTM for the first time. Self-Attentive Knowledge Tracing (SAKT)~\cite{vaswani2017attention} adds an attention mechanism to the KT model for the first time~\cite{Xia2014SequenceMemory, Liu2019Reformulating}. Context-aware attentive knowledge tracing (AKT)~\cite{Ghosh2020ContextAwareAK} model utilizes contextualized representations of practice and knowledge acquisition and combines attentional mechanisms with cognitive and psychometric models. Reconciling Cognitive Modeling with Knowledge Forgetting: A Continuous Time-aware Neural Network Approach (CT-NCM) ~\cite{Ma2022ReconcilingCM} realistically integrates the dynamics and continuity of knowledge forgetting into the modeling of student learning processes. Learning Process-consistent Knowledge Tracing (LPKT) aims to assess the student's knowledge status by modeling the student's learning process, and considers the impact of the answer time interval and answer time on the learner's knowledge status during the learning process. Monitoring Student Progress for Learning Process-consistent Knowledge Tracing (LPKT-S) is an extension of LPKT that clearly distinguishes the individual progress rate of each student. However, the above models ignore the impact of individual differences among students on the forgetting process, and lacks exploration of the causal forgetting caused by the correlation and hierarchy between knowledge concepts.

\subsection{Forgetting Curve}
In pedagogy, forgetting is a complex and universal phenomenon~\cite{10.5555/3091765.3091824}, and with time, students' knowledge proficiency may decline due to forgetting factors~\cite{Dai2020}. The Ebbinghaus forgetting curve proposes~\cite{ebbinghaus2013memory, Oiwa2014FeatureAware,nagatani2019augmenting} that students forget most quickly shortly after learning new knowledge or skills, and then the rate of forgetting gradually slows down. This also shows that when students' knowledge state tends to be stable, the proportion of forgotten knowledge will decrease\cite{10.1145/3569576}, but this is not only due to the influence of time factor but also related to knowledge structure. Trace decay theory emphasizes~\cite{loftus1985evaluating, Fu2014ComputationalCognition}that the knowledge students learn will gradually decay with time, and emphasizes the need for regular review and consolidation in the learning process. By actively reviewing and consolidating the learned knowledge, students can significantly delay the rate of forgetting, so that they can remember knowledge for a long time. This shows that, if there is a certain correlation between knowledge, then forgetting some knowledge will also lead to the forgetting of its related knowledge. However, based on this theory, through continuous review and consolidation, the knowledge will gradually transform into a stable knowledge state based on knowledge correlation when it reappears after forgetting.

\section{Preliminaries}
\subsection{Relation Definition}
\label{sec:section3.1}
Let E and C be the set of all different exercises and knowledge concepts respectively, where 
E = \{$e_{1}$, $e_{2}$, \ldots, $e_{N}$\}  \text{and} C = \{$c_{1}$, $c_{2}$, \ldots, $c_{K}$\}. 
Then problem sets and knowledge concept relation matrix can be expressed as $Q_{ij} = \{q_{e_{i}c_{j}}\}_{N \times K}$. If the exercise $e_i$ contains the knowledge concept $c_j$, $q_{e_ic_j}$ is set to 1; otherwise, $q_{e_ic_j}=0$. In addition, inspired by cognitive diagnosis and based on the directed and undirected relations between knowledge concepts, we construct a knowledge concept prerequisite relation matrix similar to \Qmatrix~\cite{liu2012data}, that is \Pmatrix, representing the directed relations between different knowledge concepts. This can be expressed as $P_{ij} = \{p_{c_{i}c_{j}}\}_{K \times K}$. If there is a leading relation between the knowledge concepts \(c_i\) and \(c_j\), then $p_{c_ic_j}=1$; otherwise, $p_{c_ic_j}=0$.

\subsection{Problem Statement} 
For students, the learning process for each time step is expressed as an interaction process, which is represented by ($e_{t}$, $a_{t}$), where $e_{t}$ represents the index of the exercise undertaken at time step $t$, and $a_{t}$ = 1 indicates a correct response, while $a_{t}$= 0 indicates an incorrect response. Knowledge tracing (KT) aims to forecast a student's response to a given exercise based on their interaction history. In a formal sense, given the interaction record in the previous T time steps denoted by R={($e_{1}$, $a_{1}$),..., ($e_{t}$, $a_{t}$),..., ($e_{T}$, $a_{T}$)}, the KT model seeks to assess the mastery level of the knowledge concept and predict the student's response to the exercise at time step T+1 ($e_{T+1}$).

\section{Methodology}
\begin{figure*}
    \centering
    \includegraphics[width=1.0 \linewidth]{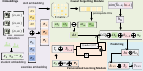}
    \caption{The main network structure of \methodname model is mainly composed of learning, forgetting and prediction modules. The learning module uses the learning gate to get the personalized learning gain, uses the directed relationship of knowledge concepts to capture the causality of forgetting process, and finally predicts the future performance of students.}
    \label{fig:network}
    \label{fig:model}
\end{figure*}

In this section, we introduce the proposed Concept-driven Personalized Forgetting knowledge tracing method (\methodname) in detail. 
First, we briefly review the baseline,\textit{i.e.}, LPKT, of our \methodname. Then, we present the framework of our \methodname model in detail.
As shown in Fig.~\ref{fig:model}, \methodname utilizes personalized knowledge acquisition and forgetting mechanisms to effectively update students' knowledge status. Simultaneously, by modeling the hierarchical relationship between knowledge concepts, it reveals the causal relationship of the forgetting process.

\subsection{Overview of LPKT}
LPKT consists of three modules at each learning step, including learning module, forgetting module and prediction module. Specifically, the learning phase mainly calculates the positive impact of learning gain by calculating the difference between the current interaction and an adjacent interaction, and models the student's knowledge status. In response to the negative impact of forgetting, the forgetting stage uses a forgetting gate to determine the degree of decay of the knowledge state over time. Then the knowledge mastery level in the learning stage and forgetting stage is used to update the knowledge status, and finally the proposed prediction module is used to predict the student's performance in the next exercise.

In the learning stage, LPKT can model the learning gains by connecting the students' previous learning embedding $l_{t-1}$ and the current learning embedding $l_{t}$ as the basic input element of LPKT. The current learning embedding $l_{t}$ consists of exercise $e_{t}$, answer $a_{t}$, and answer time $at_{t}$. LPKT considers two factors impacting learning outcomes: the interval time and the student's prior knowledge. To focus on the state of the knowledge concept related to the current exercise, LPKT multiplies the knowledge state $h_{t-1}$ with the knowledge concept vector $q_{e_{t}}$ of the current exercise to get the related knowledge state $\widetilde{h}_{t-1}$. 
\begin{equation}
{l}_{t} = \mathbf{W}_{1}^T[{e}_{t}  \oplus {a}_{t}  \oplus {at}_{t}] + {b}_{1},
\end{equation}
\begin{equation}
     \tilde{h}_{t-1} = {q}_{e_{t}} \cdot {h}_{t-1},
\end{equation}
Considering that not all learning gains directly contribute to the increasing of students' knowledge, the learning gate is designed to control students' ability to absorb knowledge and ensure a positive learning gain through linear transform. 

\begin{equation}
    {lg}_{t} = \tanh( \mathbf{W}_{2}^T[{l}_{t-1}\oplus {it}_{t} \oplus {l}_{t} \oplus \widetilde{h}_{t-1}] +{b}_{2}),
\end{equation}
\begin{equation}
    {\Gamma}_{t}^{l} = \sigma ( \mathbf{W}_{3}^T [{l}_{t-1}\oplus {it}_{t} \oplus {l}_{t} \oplus \widetilde{h}_{t-1}] + {b}_{3} ),
\end{equation}
\begin{equation}
    {LG}_{t} = {\Gamma}_{t}^{l} \cdot (({lg}_{t} + 1) /2 ),
\end{equation}

According to the forgetting curve theory, the memory of learned material declines exponentially over time. LPKT employs a forgetting gate to simulate the forgetting effect, primarily determined by the students' prior knowledge, learning intervals, and learning gains. Then, the output of the forgetting gate ${\Gamma}_{t}^{f}$ is multiplied by $h_{t-1}$ to eliminate the influence of forgetting, which can obtain the required updated knowledge state $h_{t}$ of students following the \textit{t}-th learning interaction.
\begin{equation}
    {\Gamma}_{t}^{f} = \sigma(\mathbf{W}_{4}^T[{h}_{t-1} \oplus {LG}_{t} \oplus {it}_{t}] + {b}_{4}),
\end{equation}
\begin{equation}
    {h_{t}} = \widetilde{LG}_{t} + {\Gamma}_{t}^{f} \cdot {h_{t-1}}.
\end{equation}

\subsection{Proposed Method}
\subsubsection{Personalized Learning Module}
In LPKT, there is no clear distinction between students’ individual abilities, but the learning process of each student is diﬀerent. LPKT only distinguishes diﬀerent students’ responses to the same exercises through the answer time, ignoring the modeling of students’ abilities. In order to address this limitation, we explicitly introduce a  personalized student ability $s_t$ in our methods. Specifically, as show in Eq.8, we derive this personalized student ability \(s_t  \in \mathbb{R}^{d_k} \) by combining the problem difficulty \(df_t  \in \mathbb{R}^{d_k} \), answer time \(at_t  \in \mathbb{R}^{d_k} \), correct accuracy \(ac_t \in \mathbb{R}^{d_k} \). Simultaneously, we combine the  exercise embedding $e_t$ with knowledge concepts to get the concept-perceived exercise embedding $\tilde{e}_{t}$. Then we concatenate the $\tilde{e}_{t}$, $a_t$, $s_t$ together and apply a perceptron to extract the personalized learning emdedding $\tilde{l}_{t}$. 

\begin{equation}
{s}_{t} = \alpha \otimes {df}_{t} + \beta \otimes {ac}_{t} + \mu \otimes {at}_{t},
\end{equation}
\begin{equation}
\tilde{e}_{t} = {e}_{t} \oplus {c}_{t},
\end{equation}
\begin{equation}
\widetilde{l}_{t} = \mathbf{W}_{1}^T[\tilde{e}_{t}  \oplus {a}_{t}  \oplus {s}_{t}] + {b}_{1},
\end{equation}
where $\mathbf{W}_{1} \in \mathbb{R}^{(d_{e} + d_{k} + d_{a}) \times d_{k}}$ and ${b}_{1} \in \mathbb{R}^{d_{k}}$ denote the weight matrix and its corresponding bias term. $\alpha$, $\beta$, $\mu$ are hyper-parameters. $\otimes$ and $\oplus$ denote the element-wise multiplication and concatenation operation. $d_{k}$ represents the dimension.

Considering the learning gain in LPKT requires combining the learning unit of the previous moment and the time interval for answering the exercise, we simplify this steps and further combine the student’s ability to achieve personalized learning gain as follows: 
\begin{equation}
\widetilde{lg}_{t} = \textit{tanh}(\mathbf{W}_{2}^T[\widetilde{l}_{t}\oplus \widetilde{h}_{t-1}] +{b}_{2}),
\end{equation}

where $\mathbf{W}_{2} \in \mathbb{R}^{(2d_{k})\times d_{k}}$ denotes the weight matrix, $b_{2}\in \mathbb{R}^{d_{k}}$ denotes the bias term, \textit{tanh} denotes the non-linear activation function. Then we apply a perceptron to generate a personalized learning gain gate as:
\begin{equation}
\widetilde{\Gamma}_{t}^{l} = \sigma ( \mathbf{W}_{3}^T [\widetilde{l}_{t} \oplus \widetilde{h}_{t-1} ]+ {b}_{3} )
\end{equation}

where $\mathbf{W}_{3} \in \mathbb{R}^{(2d_{k})\times d_{k}}$ denotes the weight matrix, ${b}_{3}\in \mathbb{R}^{d_{k}}$ is the bias term, $\sigma$ denotes the non-linear \textit{sigmoid} activation function.

Furthermore, we think that the mastery of the knowledge concepts contained in a certain exercise may have a positive impact on the mastery of their related knowledge concepts in the learning process. In order to capture this impact between the relevant knowledge concepts, we directly concat the knowledge state $h_{t-1}$ with the relevant knowledge concept vector $ p_{e_t}$ to obtain the degree of mastery of relevant knowledge concepts $\tilde{h}_{t-1}^{c}$ as:
\begin{equation}
\tilde{h}_{t-1}^{c} = {p}_{e_{t}} \oplus {h}_{t-1},
\end{equation}

${p}_{e_{t}} \in \mathbb {R}^{d_ {k}}$ represents the relevant knowledge concept vector, which is obtained by matching the related knowledge concepts of the current knowledge concept in \Pmatrix defined in section \ref{sec:section3.1}. By embedding the relevant knowledge concepts $\tilde{h}_{t-1}^{c}$ into the learning interaction, we can get the final knowledge acquisition $\widetilde{LG}_{t}^{c}$ as follows:
\begin{equation}
{LG}_{t}^{c} = {\widetilde\Gamma}_{t}^{l} \cdot [(\widetilde{lg}_{t} + 1) /2 ] + ||\sigma (\tilde{h}_{t}^{c})||_{2}
\end{equation}
\begin{equation}
    \widetilde{LG}_{t}^{c}  = {q}_{e_{t}} \cdot {LG}_{t}^{c}
\end{equation}

\subsubsection{Causal Forgetting Module}
The student forgetting process is similar to the learning process. In the learning process, mastering a specifc knowledge point may have a positive impact on related knowledge concepts and different students always get personalized knowledge gain in each interaction. Conversely, in the forgetting process, forgetting a knowledge point can negatively inﬂuence related knowledge points. Simultaneously, students with different cognitive abilities may also represent individual forgetting rate. Therefore, we design a personalized and causal forgetting module to capture this inﬂuence and monitor the personalized forgetting situation in this section.

\textbf{Causal forgetting mechanism}. In order to capture the causal relationship between the knowledge concept contained in the current exercise and its related knowledge concept in the forgetting process.
we design a causal forgetting mechanism in this study. 
Specifically, we first obtain the response time $ at_t $ and the interval
time $it_t$ for the current exercise, and identify the knowledge concept contained in the current exercise at the same time. Then, we use the P-matrix defined in section \ref{sec:section3.1} to match all prerequisite knowledge concepts corresponding to this knowledge concept, and further obtain the response time $at_t$ and interval time $it_t$ for exercises corresponding to these prerequisite knowledge concepts. It should be noted that there is a one-to-one correspondence between exercises and knowledge concepts, and a one-to-many relationship between knowledge concepts. So we determine the closest prerequisite knowledge concept to the
current knowledge concept by considering the time step intervals between the exercises corresponding to these
prerequisite knowledge concepts and the current exercise. For example, if the exercise $e_m$ contains the prerequisite knowledge concept $c_j$ of the knowledge concept $c_i$ in current exercise $e_t$ and the exercise $e_m$ is closest to the current exercise $e_t$, we will select knowledge concepts in the exercise $e_m$ as the closest prerequisite knowledge concept to the
current knowledge concept. Finally, as defined in Eq.16 and Eq.17, we construct a forgetting weight by calculating
the time diﬀerence between the exercises corresponding to the nearest prerequisite knowledge concept and the current exercise: 

\begin{equation}
\triangle {t}_{tm} = \lvert ({at_{t}} + {it_{t}}) - ({at_{m}} + {it_{m}}) \rvert \quad \text{where } t > m, \quad 0 \leq t, m \leq T-1
\end{equation}
\begin{equation}
w_{tm}= \delta \left (1+(exp(\triangle {t}_{tm}+\lambda ))\right)^{-1} 
\end{equation}

The hyperparameter $\lambda$ is used to adjust the oﬀset of the forgetting weight. If the knowledge concept $c_i$ in the current exercise successfully matches a prerequisite knowledge concept $c_j$ in the \Pmatrix, then $p_{ij}\neq 0$; otherwise, $p_{ij}=0$. Furthermore, we believe that forgetting a knowledge concept not only aﬀects the mastery of that specifc concept but also impacts the mastery of its related concepts. Therefore, we multiply the forgetting weight calculated based on the hierarchical relationship between knowledge concepts by the learning gain to obtain the final overall knowledge gain $LG_{t}^{l}$:
\begin{equation}
    {w}_{f} = \left\{
\begin{aligned}
    &1 \quad &\text{if } {p}_{ij} = 0 \\
    &{w}_{tm} \quad &\text{if } {p}_{ij} \neq 0
\end{aligned}
\right.
\end{equation}

\begin{equation}
{LG}_{t}^{l} = w_{f} \cdot \widetilde{LG}_{t}^{c}
\end{equation}

Given that students’ memory of learning materials tends to decline over time. However, when students encounter exercises similar to those they have previously encountered, they may undergo a process of knowledge reactivation, known as the review process, which aligns with the phenomena of long-term forgetting and short-term memory. We call it the forgetting-review mechanism of the learning process. We simulate this mechanism and analyze the changes in short-term knowledge mastery by computing the similarity of k adjacent knowledge states, as shown in Eq.20. And then, we apply this similarity to generate a related knowledge state $h_k$, as defined in Eq.21, at the current moment. This related knowledge state $h_k$ will be further applied in the Eq.22 to updata the forgetting gate.
\begin{equation}
    {m}_{k}= \text{sim}(\tilde{h}_{t-k} ,  {h}_{t-1})
\end{equation}
\begin{equation}
    {h}_{k} =\sum_{k=1}^{T} \text{softmax}(m_k) \cdot {h}_{t-1} 
\end{equation}

\textbf{Personalized forgetting mechanism}. Individual diﬀerences can lead to variations in students’ knowledge acquisition and forgetting rates when facing the same sequence of exercises. In order to calculate students' knowledge states more accurately, we incorporate students' personalized capabilities into the modeling of the forgetting process. By introducing the dynamic mechanism of the forgetting gate ${\widetilde\Gamma}_{t}^{f}$, we integrate students' personalized capabilities $s_t$, their previous knowledge state $h_{t-1}$, their knowledge mastery ${LG}_{t}^{l}$, the time interval between exercises $it_t$, and their short-term memory degree $h_k$ to simulate knowledge forgetting, thereby enhancing our ability to predict knowledge states.
\begin{equation}
\widetilde{\Gamma}_{t}^{f} = \sigma(\mathbf{W}_{4}^T[{h}_{t-1} \oplus {LG}_{t}^{l} \oplus {it}_{t} \oplus {s}_{t} \oplus {h}_{k}] + {b}_{4})
\end{equation}
Where $\mathbf{W}_{4}\in \mathbb{R}^{(5d_{k}\times d_{k})}$ denotes the weight matrix, ${b}_{4}\in \mathbb{R}^{d_{k}}$ denotes the bias term, $\sigma$ denotes the non-linear \textit{sigmoid} activation function. Similar to LPKT, we multiply the forgetting factor $\widetilde{\Gamma}_{t}^{f}$ by the previous state of knowledge ${h_{t-1}}$ to eliminate the effect of forgetting. Therefore, after completing the $t$-th learning interaction, the student's knowledge status ${h_{t}}$ is updated as follows:

\begin{equation}
{h_{t}} = \widetilde{LG}_{t}^{c} + {\widetilde\Gamma}_{t}^{f} \cdot {h_{t-1}}
\end{equation}

\subsubsection{Predicting Module}
The prediction layer combines the temporal characteristics of the KT model and makes full use of students' learning history and interaction sequences to predict their performance in the next exercise. Specifically, we predict their performance in the next exercise ${e_{t+1}}$ based on their knowledge state ${\widetilde{h}_{t}}$ after the $t$-th learning interaction. We first concatenate the knowledge state embedding $\widetilde{h}_{t}$ with the concept-perceived exercise embedding $\widetilde{e}_{t+1}$, and then project them onto the output layer through a fully connected network with sigmoid activation: 
\begin{equation}
{y_{t+1}} = \sigma(\mathbf{W}_{5}^T[ \widetilde{e}_{t+1} \oplus {\widetilde{h}_{t}}] + {b}_{5})
\end{equation}
Where $\mathbf{W}_{5}\in \mathbb{R}^{(2d_{k}\times d_{k})}$ denotes the weight matrix, and ${b}_{5}\in \mathbb{R}^{d_{k}}$ represents the bias term. The output $y_{t+1}$ serves to predict the student's performance in the next exercise $e_{t+1}$. Additionally, we set a threshold to determine whether the student can answer $e_{t+1}$ correctly. Specifically, if $y_{t+1}$ surpasses the threshold, the answer is a correct response; otherwise, the answer is deemed incorrect.

\subsubsection{Objective Function}
To optimize all parameters in \methodname, we employ the cross-entropy log loss between the predicted values $y$ and the actual answers $a$ as the objective function:
\begin{equation}
    \mathbb{L}(\theta)  = -\sum_{t=1}^{T} \left(a_t \log({y}_{t}) + (1 - a_t) \log(1 - {y}_{t})\right) + \lambda_{\theta} \|\theta\|^2
\end{equation}
Where $\theta$ represents all parameters in \methodname model and $\lambda_{\theta}$ serves as the regularization hyper-parameter. We employ the Adam optimizer to minimize the objective function on mini-batches. More detailed information regarding the experimental settings are provided in the subsequent sections.

\section{EXPERIMENTS}
\subsection{Experimental Setup}
In this section, we provide the detailed information of the evaluation datasets, training details and other competitive methods for comparison.
\subsubsection{Dataset Description}
In our experiments, we utilize three real-world public datasets to assess the performance of \methodname, \textit{i.e.}, ASSISTments 2012, ASSISTments Challenge, EdNet-KT1. The basic statistics for these datasets are summarized in Table~\ref{tab:dataset}.

\sloppy
\begin{itemize}
\item \textbf{ASSISTments 2012} \textbf{(ASSIST2012)} \textsuperscript{1} is the largest version of the dataset, collects between September 2012 and October 2013. It includes 17,999 exercises answered by 46,674 students, with a total of 6,123,270 interactions~\cite{10.1007/s11257-009-9063-7}. To focus on the 265 knowledge points, we filter the dataset to include only relevant exercises.
\item \textbf{ASSISTments Challenge} \textbf{(ASSISTChall)} \textsuperscript{2} is a competition organized by the Assistants online Tutoring System to promote data mining and machine learning research in the field of education. The dataset contains the learning behavior and performance data of the students in the assistments system.
\item \textbf{EdNet-KT1} \textsuperscript{3} dataset contains all student system interactions collected over two years with more than 780,000 users in South Korea~\cite{choi2020ednet}. Each student generates an average of 441.20 interactions. It offers large-scale real-world Intelligent Assisted Instruction system (ITS) data and includes 13,169 exercises, 1,021 lectures, and labels for 293 concepts. When dealing with a problem with multiple associated knowledge concepts, only the first one is chosen.
\end{itemize}

\begin{table}[h]
\centering
\caption{Statistics of all datasets.}
\begin{tabularx}{0.7\textwidth}{@{\extracolsep{\fill}}lccc}
\hline
\multirow{2}{*}{Statistics}  &                                & \multicolumn{1}{c}{Datasets}    &                               \\ \cline{2-4} 
                             & \multicolumn{1}{c}{ASSIST2012} & \multicolumn{1}{c}{ASSISTchall} & \multicolumn{1}{c}{EdNet-KT1} \\ \hline
Students & 29,018                         & 1,709                           & 784,309                       \\
Exercises                    & 53,091                         & 3,162                           & 12,372                        \\
Concepts                     & 265                            & 102                             & 141                           \\
Answer Time                  & 26,747                         & 1,326                           & 9,292                         \\
Interval Time                & 29,748                         & 2,839                           & 41,830                        \\
Avg. Length                   & 93.45                          & 551.68                          & 125.45                        \\ \hline
\end{tabularx}
\label{tab:dataset}
\end{table}

\footnotetext[1] {https://sites.google.com/site/assistmentsdata/home/2012-13-school-data-withaffect}
\footnotetext[2] {https://sites.google.com/view/assistmentsdatamining/dataset}
\footnotetext[3] {http://ednet-leaderboard.s3-website-ap-northeast-1.amazonaws.com/}

\subsubsection{Training Details}
First, we sort students’ learning records according to the timestamp of their answers. We then set all input sequences to a fixed length based on the average sequence length of the dataset. Specifically, in the ASSIST2012 and EdNet-KT1 datasets, we set the fixed length of the input sequence to 100, while in the ASSISTchall dataset, the fixed length of the input sequence is set to 500. For sequences longer than a fixed length, we cut them into multiple unique subsequences, each of which is a fixed length. For shorter sequences of fixed length, we pad them with zero vectors so that they reach fixed length.

We perform standard 5-fold cross-validation on all datasets. In each fold, 80$\%$ of the students are divided into the training set (80$\%$) and the verification set (20$\%$), the rest 20$\%$ are used as the testing set. During the training process, we randomly initialize all parameters, using uniform distribution~\cite{Glorot2010UnderstandingTD}. All hyper-parameters are learned on the training set, and the model that performs best on the validation set is selected for testing set evaluation. In \methodname, we added a dropout layer with a dropout rate of 0.2 to avoid overfitting. In our implementation, the parameters $d_{k}$ and $d_{e}$ are set to 128 and $d_{a}$ to 50. For the small positive $\gamma$ in the enhanced Q matrix, we set it to 0.03, and the relationship strength coefficient $\rho$ in the \Pmatrix is set to 0.03, the learning rate is set to $3\times 10^{-3}$, and the batch-size is set to 128. For a fair comparison, all models are trained on a cluster of Linux servers with NVIDIA 3090 GPUs.

\subsubsection{Other Competitive Methods}
To verify the effectiveness of \methodname, we compare our model with several previous methods.
\begin{itemize}
\item \textbf{DKT} applies deep learning to KT for the first time. It takes the learning sequence as input to the RNN or its variant LSTM~\cite{10.1162/neco.1997.9.8.1735, Yang2018FaultDiagnosis} and represents the student's state of knowledge through hidden states.
\item \textbf{DKT+} is an extended and improved version of DKT that aims to solve two major problems in DKT. First, DKT cannot efficiently reconstruct unobserved input data. Second, the prediction performance of DKT is inconsistent between different time steps~\cite{Yeung2018AddressingTP}. By enhancing the original DKT loss function with two additional regularization terms, DKT+ attempts to overcome these limitations and improve the performance and application range of the model.
\item \textbf{DKVMN} is a model that introduces memory-enhancing neural networks into knowledge tracing. The model uses the learning mechanism of key-value pair memory to capture the knowledge state of students~\cite{10.1145/3038912.3052580} and is applied to the student knowledge model and prediction model.
\item \textbf{SAKT} models the student's knowledge state and learning progress by taking the student's learning interaction sequence~\cite{Pandey2019ASA} as input and using self-attention mechanisms~\cite{vie2019knowledge}.
\item \textbf{AKT} is a context-aware attentive knowledge tracing model~\cite{Ghosh2020ContextAwareAK}, utilizes dual self-attentive encoders for exercises and responses.  The knowledge retriever employs attention to retrieve past knowledge relevant to the current exercise.
\item \textbf{DTransformer} establishes an architecture~\cite{10.1145/3543507.3583255} from the problem level to the knowledge level, explicitly diagnosing learners' proficiency based on the mastery of each problem. It utilizes contrastive learning to maintain the stability of diagnosing knowledge states.
\item \textbf{CT-NCM} integrates dynamic forgetting~\cite{Ma2022ReconcilingCM} into students' learning process, effectively models knowledge learning and forgetting, and distinguishes positive and negative reactions.
\item \textbf{LPKT} displays the modeling user's knowledge state and uses the embedding of exercises. The knowledge points contained in exercises are used to select the corresponding exercise records~\cite{10.1145/3447548.3467237}. It reflects the consistency of students' changing knowledge state and learning process.
\item \textbf{LPKT-S} considers that students generally have different progress rates~\cite{9950313}, a student embed containing student-specific progress rates is introduced, extending LPKT to LPKT-S.
\sloppy
\end{itemize}
\begin{table*}[h]
\caption{Results of comparison methods on students' performance prediction.}
\resizebox{1.01\linewidth}{!}{
\renewcommand{\arraystretch}{1.4}
\begin{tabular}{@{}c|llll|llll|llll@{}}
\toprule
\multirow{2}{*}{Method} & \multicolumn{4}{c|}{ASSIST2012}                                                                                   & \multicolumn{4}{c|}{ASSISTchall}                                                                                  & \multicolumn{4}{c}{EdNet-KT1}                                                                                     \\ \cmidrule(l){2-13} 
                        & \multicolumn{1}{c}{RMSE}   & \multicolumn{1}{c}{AUC}    & \multicolumn{1}{c}{ACC}    & \multicolumn{1}{c|}{$r^2$} & \multicolumn{1}{c}{RMSE}   & \multicolumn{1}{c}{AUC}    & \multicolumn{1}{c}{ACC}    & \multicolumn{1}{c|}{$r^2$} & \multicolumn{1}{c}{RMSE}   & \multicolumn{1}{c}{AUC}     & \multicolumn{1}{c}{ACC}    & \multicolumn{1}{c}{$r^2$} \\ \midrule
DKT~\cite{10.5555/2969239.2969296}                    & \multicolumn{1}{l}{0.4241} & \multicolumn{1}{l}{0.7289} & \multicolumn{1}{l}{0.7360} & 0.1468                  & \multicolumn{1}{l}{0.4471} & \multicolumn{1}{l}{0.7213} & \multicolumn{1}{l}{0.6907} & 0.1425                  & \multicolumn{1}{l}{0.4508} & \multicolumn{1}{l}{0.6836}  & \multicolumn{1}{l}{0.6889} & 0.1008\\
DKT+~\cite{Yeung2018AddressingTP}                 & \multicolumn{1}{l}{0.4239} & \multicolumn{1}{l}{0.7295} & \multicolumn{1}{l}{0.7254} & 0.1497                  & \multicolumn{1}{l}{0.4502} & \multicolumn{1}{l}{0.7101} & \multicolumn{1}{l}{0.6842} & 0.1308                  & \multicolumn{1}{l}{0.4601} & \multicolumn{1}{l}{0.6429}  & \multicolumn{1}{l}{0.6733} & 0.0635                 \\
DKVMN~\cite{10.1145/3038912.3052580}                  & \multicolumn{1}{l}{0.4261} & \multicolumn{1}{l}{0.7228} & \multicolumn{1}{l}{0.7329} & 0.1398                  & \multicolumn{1}{l}{0.4503} & \multicolumn{1}{l}{0.7108} & \multicolumn{1}{l}{0.6842} & 0.1302                  & \multicolumn{1}{l}{0.4538} & \multicolumn{1}{l}{0.6741}  & \multicolumn{1}{l}{0.6843} & 0.0913                \\
SAKT~\cite{Pandey2019ASA}                   & \multicolumn{1}{l}{0.4258} & \multicolumn{1}{l}{0.7233} & \multicolumn{1}{l}{0.7339} & 0.1403                  & \multicolumn{1}{l}{0.4626} & \multicolumn{1}{l}{0.6605} & \multicolumn{1}{l}{0.6694} & 0.0822                  & \multicolumn{1}{l}{0.4524} & \multicolumn{1}{l}{0.6794}  & \multicolumn{1}{l}{0.6862} & 0.0964                 \\
DTransformer~\cite{10.1145/3543507.3583255}                & \multicolumn{1}{l}{0.4118} & \multicolumn{1}{l}{0.7698} & \multicolumn{1}{l}{0.7509} & 0.2004                  & \multicolumn{1}{l}{0.4371} & \multicolumn{1}{l}{0.7506} & \multicolumn{1}{l}{0.7078} & 0.1791                  & \multicolumn{1}{l}{0.4291} & \multicolumn{1}{l}{0.7553}  & \multicolumn{1}{l}{0.7089} & 0.1837                 \\
AKT~\cite{Ghosh2020ContextAwareAK}                & \multicolumn{1}{l}{0.4121} & \multicolumn{1}{l}{0.7706} & \multicolumn{1}{l}{0.7515} & 0.2004                  & \multicolumn{1}{l}{0.4364} & \multicolumn{1}{l}{0.7501} & \multicolumn{1}{l}{0.7080} & 0.1801                  & \multicolumn{1}{l}{0.4297} & \multicolumn{1}{l}{0.7557}  & \multicolumn{1}{l}{0.7083} & 0.1842                 \\
LPKT~\cite{10.1145/3447548.3467237}                  & \multicolumn{1}{l}{0.4089} & \multicolumn{1}{l}{0.7740} & \multicolumn{1}{l}{0.7551} & 0.2145                  & \multicolumn{1}{l}{0.4179} & \multicolumn{1}{l}{0.7939} & \multicolumn{1}{l}{0.7385} & 0.2491                  & \multicolumn{1}{l}{0.4290} & \multicolumn{1}{l}{0.7721} & \multicolumn{1}{l}{0.7106} & 0.2195                 \\
LPKT-S~\cite{9950313}                & \multicolumn{1}{l}{0.4065} & \multicolumn{1}{l}{0.7803} & \multicolumn{1}{l}{0.7584} & 0.2004                  & \multicolumn{1}{l}{0.4160} & \multicolumn{1}{l}{0.7979} & \multicolumn{1}{l}{0.7420} & 0.2558                  & \multicolumn{1}{l}{0.4263} & \multicolumn{1}{l}{0.7801}  & \multicolumn{1}{l}{0.7158} & 0.2230                 \\
CT-NCM~\cite{Ma2022ReconcilingCM}                & \multicolumn{1}{l}{0.4013} & \multicolumn{1}{l}{0.7945} & \multicolumn{1}{l}{0.7609} & 0.2386                  & \multicolumn{1}{l}{0.4096} & \multicolumn{1}{l}{0.8166} & \multicolumn{1}{l}{0.7430} & 0.2893                  & \multicolumn{1}{l}{0.4293} & \multicolumn{1}{l}{0.7691}  & \multicolumn{1}{l}{0.7271} & 0.2183                 \\
\methodname           & \multicolumn{1}{l}{$\textbf{0.3990}$} & \multicolumn{1}{l}{$\textbf{0.8026}$} & \multicolumn{1}{l}{$\textbf{0.7665}$} & {$\textbf{0.2524}$}                  & \multicolumn{1}{l}{$\textbf{0.4075}$} & \multicolumn{1}{l}{$\textbf{0.8206}$} & \multicolumn{1}{l}{$\textbf{0.7495}$} & {$\textbf{0.2921}$}                  & \multicolumn{1}{l}{$\textbf{0.4230}$} & \multicolumn{1}{l}{$\textbf{0.7980}$}  & \multicolumn{1}{l}{$\textbf{0.7295}$} & {$\textbf{0.2640}$}                 \\ \bottomrule
\end{tabular}
}
\label{tab:res}
\end{table*}
\subsection{Experimental Results}
In this section, we conducted several experiments to illustrate the interpretability of the model from different perspectives. Firstly, our model outperforms other competitive methods in predicting student performance. Secondly, we effectively consider students' personalized abilities, which align better with real-world application scenarios. Finally, we model the causal relationship of the forgetting process by considering the predecessor and successor relationships of knowledge concepts, and incorporate students' abilities to update knowledge states.
\subsubsection{Main results}
As shown in Table~\ref{tab:res}, \methodname shows varying degrees of improvement on the three datasets compared with other models, indicating that the learning and forgetting behaviors emphasized by \methodname are effective in knowledge tracing modeling.

DKT uses the latent vector of the LSTM model to model the overall knowledge status of students. It does not establish the connection between questions and knowledge concepts and explore the potential relationships between knowledge concepts, and cannot capture each student's mastery of knowledge concepts. Therefore, the predictive performance of DKT is lower than \methodname on all three datasets. 
Both DKVMN and \methodname can simulate students' mastery of various knowledge concepts, but DKVMN ignores the forgetting behavior during the learning process. Therefore, \methodname outperforms DKVMN in predictive performance. This also shows that modeling knowledge concepts can better capture changes in knowledge status and more accurately calculate students' mastery of knowledge concepts. 
SAKT and AKT both consider the decay of memory over time but do not account for individual student differences and other factors that may influence the forgetting process. 
Compared with other models, the LPKT and LPKT-S methods comprehensively model learning and forgetting behaviors in the learning process, but LPKT does not emphasize the individual differences of students. 
Although LPKT-S distinguishes individual learning progress rates, it does not capture the causal relationships present in the forgetting process.
\methodname emphasizes the impact of students' cognitive abilities in the learning and forgetting process, deeply explores the causal forgetting caused by the relationship between knowledge concepts, it can be observed that \methodname significantly outperforms the baseline model LPKT on the EdNet-KT1 dataset (\textit{i.e.}, improves the AUC by 2.59$\%$), indicating a more comprehensive modeling of both the learning and forgetting processes.
CT-NCM, similar to \methodname, is designed for modeling the forgetting process. CT-NCM adeptly incorporates the continuous and dynamic forgetting behavior into the modeling of students' learning processes, discerning the influence of both positive and negative responses on their knowledge states. However, it omits the analysis of hierarchical relationships among knowledge concepts and neglects the impact of individual differences on knowledge states. We observed that \methodname outperforms the CT-NCM method on the EdNet-KT1 dataset (\textit{i.e.}, improves the AUC by 2.89$\%$), indicating that our modeling approach to the forgetting process is closer to real-world scenarios.
\subsubsection{Students’ mastery of knowledge points}
\begin{figure}[h]
    \centering
    \includegraphics[width=1.0 \linewidth]{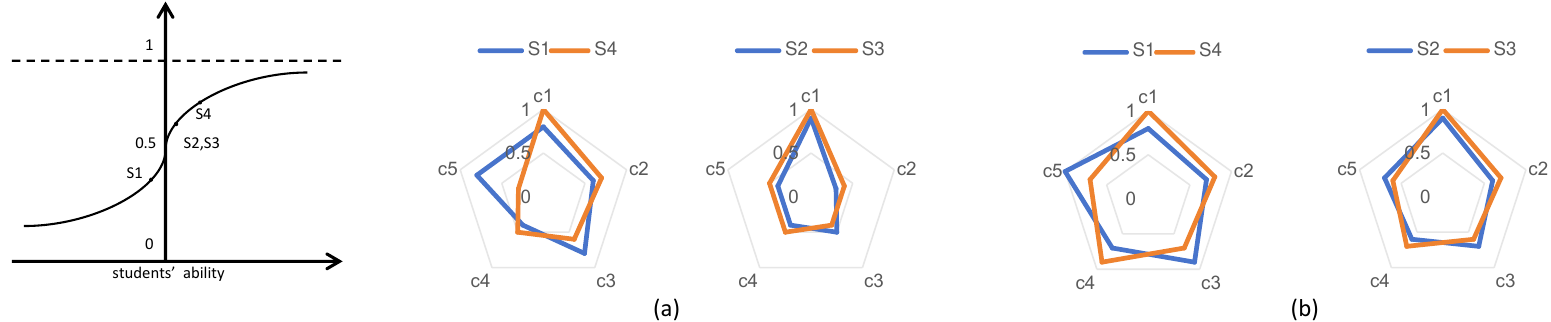}
    \caption{Illustration of the mastery level of student knowledge points. On the left side are the individual differences in students' abilities, while on the right side are the mastery levels of knowledge points after students undergo a consistent answering process. (a) and (b)  represent the initial ability distribution of students and the degree of mastery of knowledge concepts after the same answering sequence respectively. }
    \label{fig:network}
    \label{fig:per}
\end{figure}

The personalized abilities of students play a crucial role in both the learning and forgetting stages. Traditional models rely on the random initialization of student abilities. As shown in Figure~\ref{fig:per} (a) and (b), Student2 (S2) and Student3 (S3) exhibit similar initial mastery levels of the knowledge concept $c_5$, demonstrating relatively consistent abilities. After completing the same sequence of exercises, there is no difference in the final mastery level of the knowledge concept $c_5$. However, for Student1 (S1) and Student4 (S4), due to different initial mastery levels of the knowledge concept $c_4$, there are differences in their abilities. Consequently, there are differences in the mastery levels after learning. This indicates that despite making similar responses to the same exercises, students with different levels of ability obtain different learning outcomes, leading to differences in the final mastery level of knowledge.

\subsubsection{Visualization of Causal Forgetting}
\begin{figure}[h]
  \begin{minipage}[b]{0.5\textwidth}
    \centering
    \includegraphics[width=1.08\linewidth]{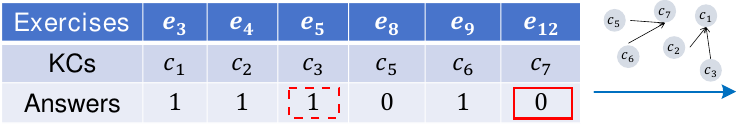}
    \vspace{0.01em}
  \end{minipage}
  \hfill
  \begin{minipage}[b]{0.4\textwidth}
  \hspace{-1.05cm}
    \centering
    \includegraphics[width=1.08\linewidth]{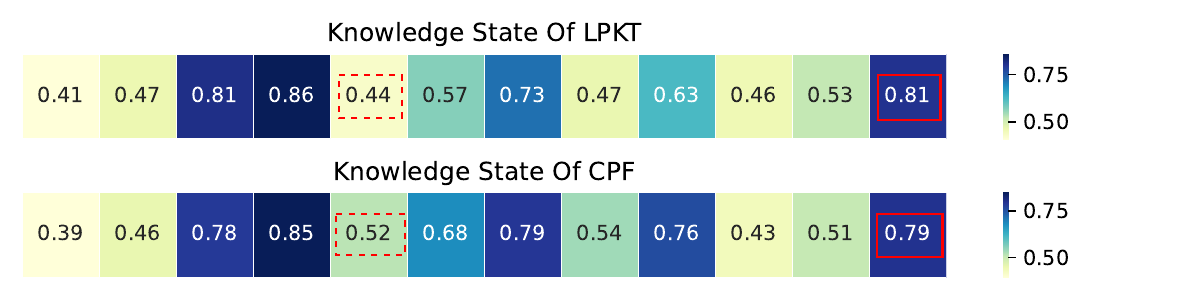}
    \vspace{0.95em}
  \end{minipage}
  \vspace{-0.4em}
  \caption{The evolution of students' knowledge state. The left side represents the students' answer order and results, and the right side describes the updating of the knowledge state due to the knowledge concept relationship.}
  \label{fig:vis}
  \vspace{-1.5em}
\end{figure}

In \methodname, we employ a novel forgetting strategy by using a matrix to represent the relationships between knowledge concepts and their prerequisites. 
During the modeling of the forgetting process, we determine the forgetting weight by calculating the closest prerequisite knowledge concept to the current one. Simultaneously, we also observe the phenomenon of memory enhancement, whereby students may forget certain knowledge concepts, but upon encountering similar exercises shortly thereafter, they reinforce their memory of these concepts through the review process.
Fig.~\ref{fig:vis} illustrates the changes in knowledge states for \methodname and LPKT. 
In the learning stage, the acquisition of prerequisite knowledge concepts contributes to subsequent knowledge concepts, while in the forgetting stage, the forgetting of subsequent knowledge concepts will have an impact on the mastery of prerequisite knowledge concepts.
For instance, $c_{1}$ and $c_{2}$, serving as prerequisites for $c_{3}$, result in a higher knowledge state in the answer $e_{5}$ compared to the baseline. In the forgetting process, when a knowledge concept that has already forgotten another encounters a similar type of exercise, the forgetting degree for subsequent knowledge concepts may exceed simple time decay. 
$c_{5}$ and $c_{6}$ as prerequisites for $c_{7}$ exhibit a weaker improvement in knowledge state when answering $e_{12}$ due to a certain level of forgetting of prerequisite knowledge concepts. 

\subsubsection{Ablation Study}
In this section, we conduct ablation experiments to demonstrate the efficacy of each module and the effect of parameter in \methodname. 
\begin{itemize}
    \item \textcolor{black}{\bfseries \methodname (P)} does not consider forgetting at the conceptual level of knowledge.
    \item \textcolor{black}{\bfseries \methodname (FP)} does not consider the impact of forgetting on learning throughout the interaction.
    \item \textcolor{black}{\bfseries \methodname (L)} does not consider the learning process of students for specific knowledge concepts and overall knowledge.
    \item \textcolor{black}{\bfseries \methodname (I)} does not take into account individualized differences in students' abilities during the learning process.
\end{itemize}
\begin{table}[ht]
\caption{Ablation experimental results on ASSIST2012.}
\centering

\begin{minipage}{0.50\textwidth}
  \centering
  \resizebox{\textwidth}{!}{%
    \begin{tabular}{lcccc}
      \toprule
      Model    & P-matrix & Personalization & AUC    & ACC  \\
      \midrule
      \methodname(P)  & \ding{55} & \ding{51} & 0.7821 & 0.7562  \\
      \methodname(I)  & \ding{51} & \ding{55} & 0.8008 & 0.7658 \\
      \methodname     & \ding{51} & \ding{51} & $\textbf{0.8026}$ & $\textbf{0.7665}$  \\
      \bottomrule
    \end{tabular}%
    }
    
\end{minipage}%
\hfill
\begin{minipage}{0.47\textwidth}
  \centering
  \resizebox{\textwidth}{!}{%
    \begin{tabular}{lcccc}
      \toprule
      Model    & Learning & Forgetting & AUC    & ACC  \\
      \midrule
      \methodname(L)  & \ding{55} & \ding{51} & 0.7905 & 0.7593  \\
      \methodname(FP) & \ding{55} & \ding{55} & 0.7725 & 0.7538 \\
      \methodname     & \ding{51} & \ding{51} &  $\textbf{0.8026}$ & $\textbf{0.7665}$  \\
      \bottomrule
    \end{tabular}%
  
  }
\end{minipage}

\label{tab:ablation}
\end{table}
Apart from the settings mentioned above, the remaining components and experimental configurations of the model remain unchanged. The results in Table~\ref{tab:ablation} reveal some interesting findings. Firstly, the commonly observed phenomenon of forgetting plays a crucial role in the learning process, and ignoring the relationships between knowledge concepts when modeling the forgetting process can result in poorer predictive outcomes. Secondly, because our model explores the causal relationships of the forgetting process in more detail, ignoring the entire forgetting process would greatly diminish predictive results. This also indicates that our designed forgetting process more comprehensively considers factors influencing forgetting, better simulating real-world application scenarios. Thirdly, if the importance of knowledge concepts is not considered in modeling the learning process, there will also be a certain degree of decline in predictive results. Fourthly, neglecting student personalization would also decrease the predictive accuracy of the model.
\subsection{Experimental Analysis}
In this section, we analyze the distribution of relationships between knowledge concepts in the dataset, discuss the reasons for selecting directed relationships, and explore parameter sensitivity.
\subsubsection{Construction of Knowledge Concept Relation}
Inspired by Relation Map Driven Cognitive Diagnosis (RCD), we introduce the following matrix to build dependencies between concepts~\cite{10.1145/3350546.3352513,wang2024studentaffect}. The first is the Answer Matrix, denoted by $A$. From the exercise record, we calculate the matrix $A$, where $A_{ij} $ represents the number of times the concept $j$ is answered correctly immediately after the concept $i$ is answered correctly, and $A_{i,j}= \frac {n_{i,j}}{\sum_{k} n_{i,k}}$ ( if $i\neq j$), else, it is 0.  
In addition, we have a Transition Matrix, denoted by $T$, which is a binary matrix~\cite{10.1145/3404835.3462932,wang2023selfsupervised}. Where $T_{i,j} = 1$ indicates that there is an edge from concept $i$ to concept $j$. To obtain the transfer matrix $T$, we first calculate the normalized matrix of matrix $A$, denoted as \( \widetilde{T} \).
Specifically, $\widetilde{T}_{ij} = \frac{{A_{ij}-min(A)}}{max(A)-min(A)}$ represents the probability that some educational relationship exists between concept $i$ and concept $j$. 
Then, we determine the relations by $T_{i,j} = 1$ if $\widetilde{T}_{ij}$ $>$ $threshold$.
we set $threshold$ as third power of the average value of matrix $T$.
If $T_{i,j} = 1$ but $T_{j,i} \neq 1$, then the concept $j$ is a successor to the concept $i$. The transition matrix $T$ calculated in this way can represent the hierarchical relationship between concepts.
\begin{figure}[h]
  \begin{minipage}[b]{0.3\textwidth}
    \centering
    \includegraphics[width=0.9\linewidth]{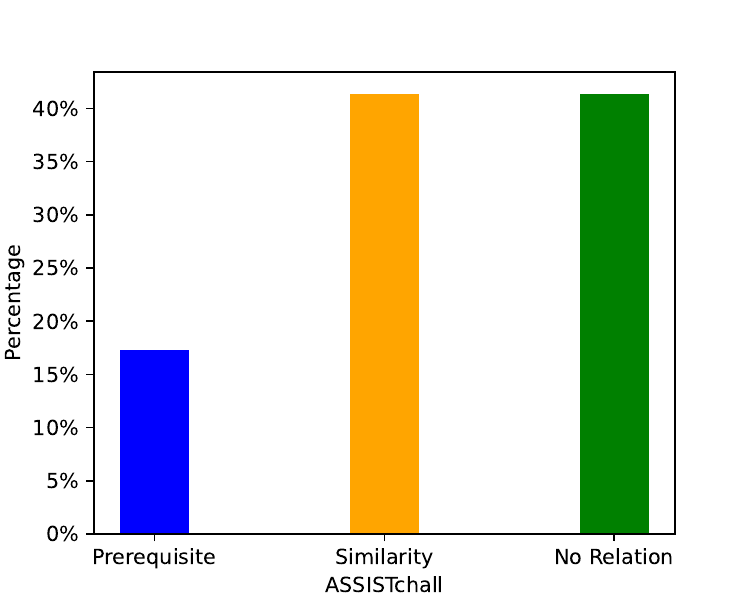}
  \end{minipage}
  \hfill
  \begin{minipage}[b]{0.3\textwidth}
    \centering
    \includegraphics[width=0.9\linewidth]{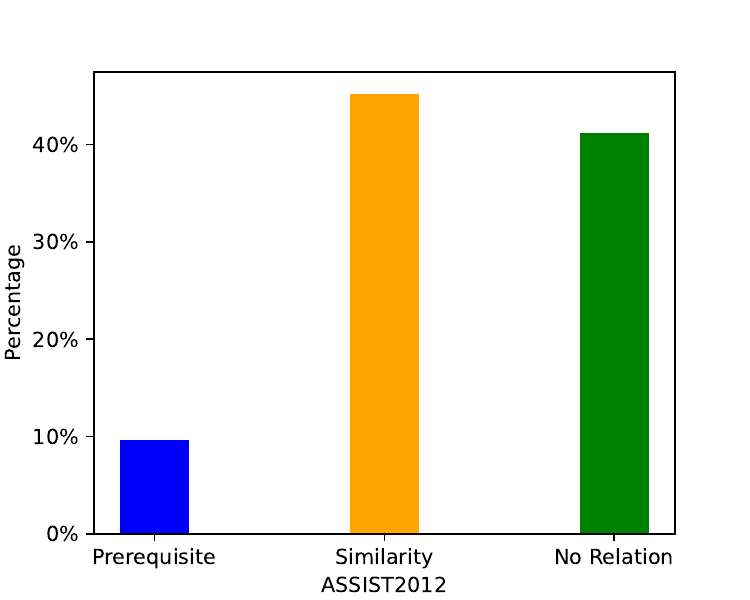}
  \end{minipage}
  \hfill
  \begin{minipage}[b]{0.3\textwidth}
    \centering
    \includegraphics[width=0.9\linewidth]{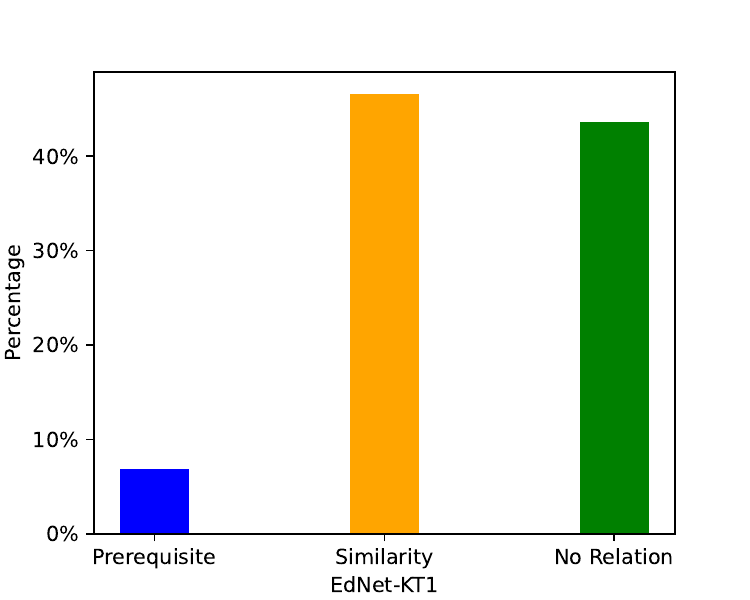}
  \end{minipage}
  \caption{Comparison of exercise-concept correlation study}
  \label{fig:exercise_concep}
  \vspace{-1.5em}
\end{figure}
\subsubsection{Knowledge Concept Relationship Analysis}
Fig.\ref{fig:exercise_concep} depicts three types of relationships between knowledge concepts in three datasets: (1) the $c_{i}$ concept is similar to the $c_{j}$ concept; (2) The concept of $c_{i}$ is the prerequisite knowledge concept of $c_{j}$; (3) There is no clear relationship between the $c_{i}$ concept and the $c_{j}$ concept. 
The distribution trends of the three datasets are similar, with the lowest proportion of prerequisite relationships, and the higher proportions of similar relationships and no clear relationships. We rely on these relationships to measure the degree of forgetting of relevant knowledge concepts. Due to the large proportion of similar relationships, choosing to use similar relationships for weighting may have an impact on many exercises in the datasets, which will cause the weighting effect to be insignificant. Therefore, we decided to use prerequisite relationships to weight the forgetting process. From the results in Table~\ref{tab:res}, it can also be observed that the experimental results of ASSISTchall show relatively minor improvements, which reinforces the validity of our choice of prerequisite relationships as the weighting method.
\begin{figure}[h]
  \begin{minipage}[b]{0.23\textwidth}
    \centering
    \includegraphics[width=\linewidth]{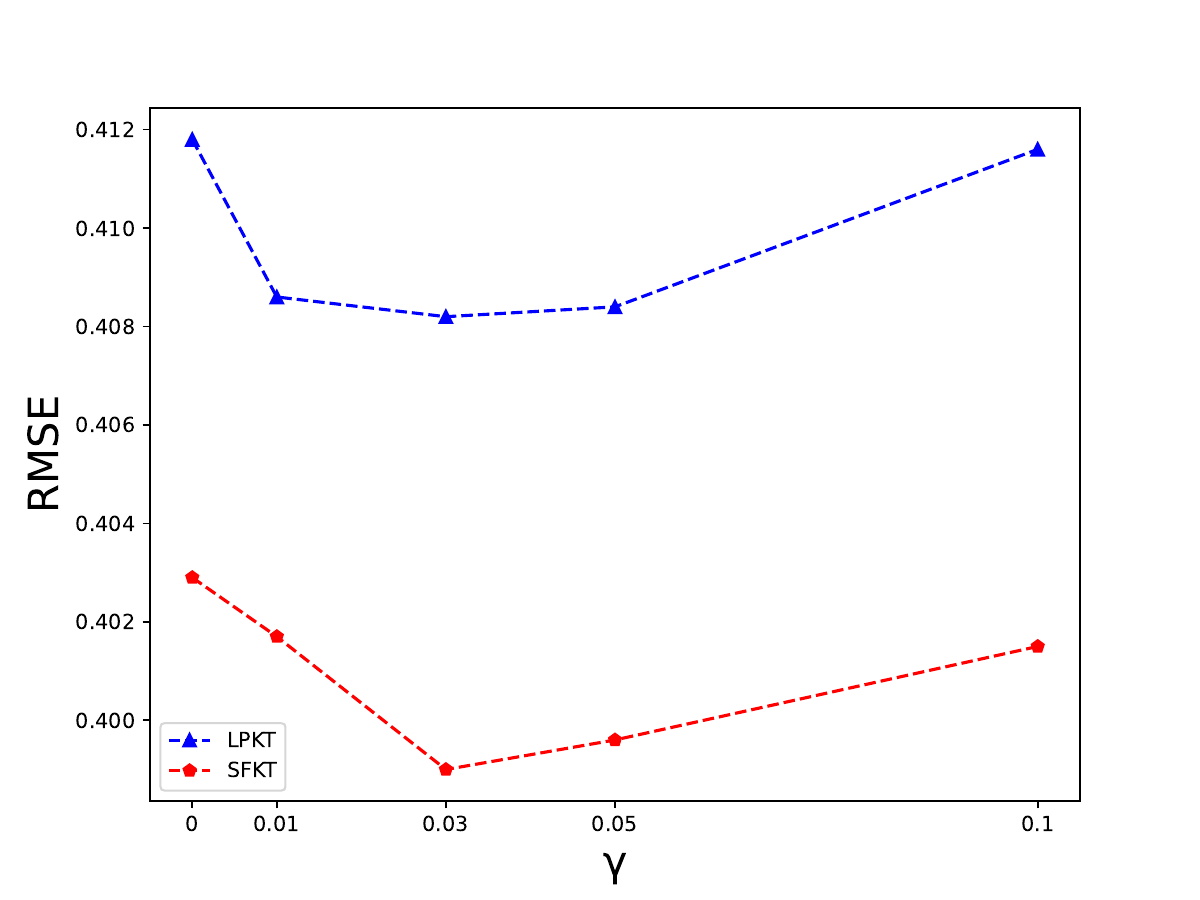}
  \end{minipage}
  \hfill
  \begin{minipage}[b]{0.23\textwidth}
    \centering
    \includegraphics[width=\linewidth]{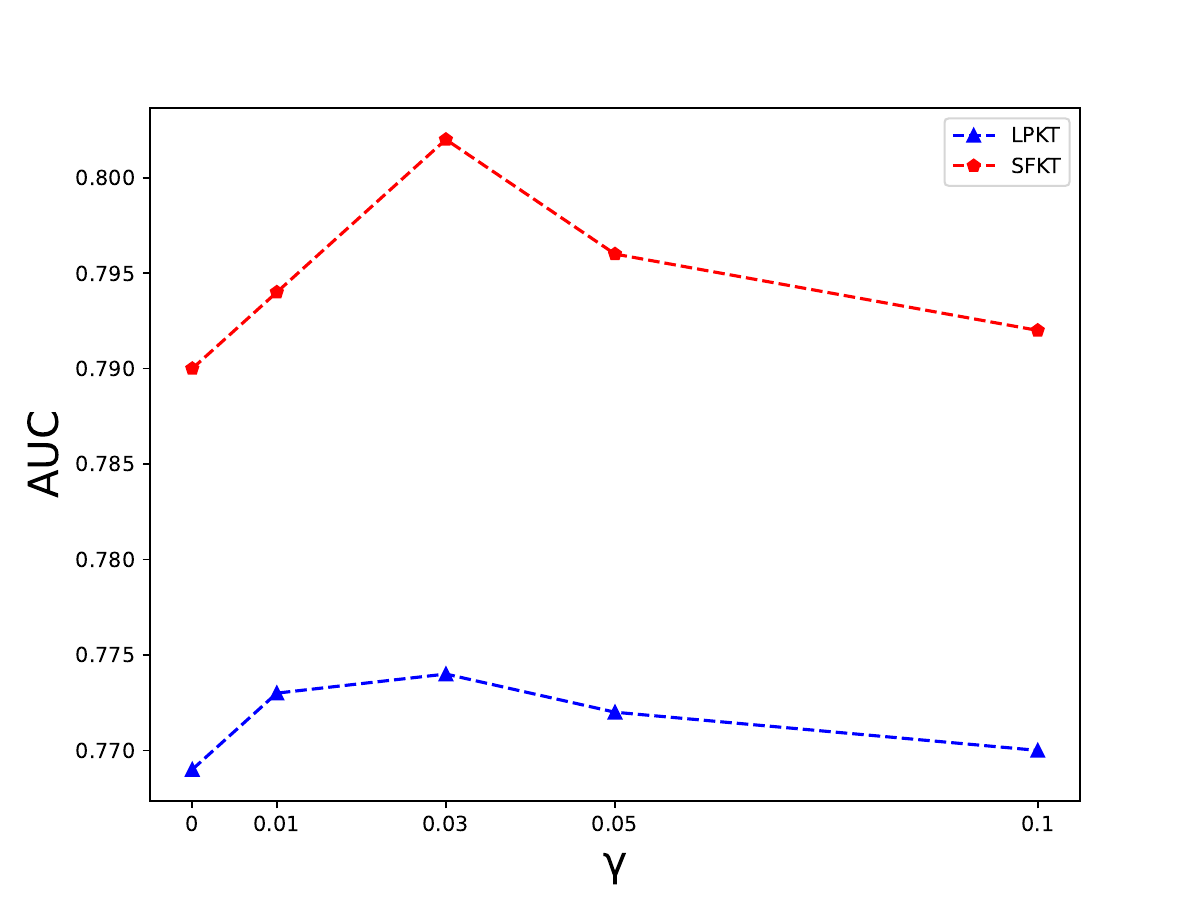}
  \end{minipage}
  \hfill
  \begin{minipage}[b]{0.23\textwidth}
    \centering
    \includegraphics[width=\linewidth]{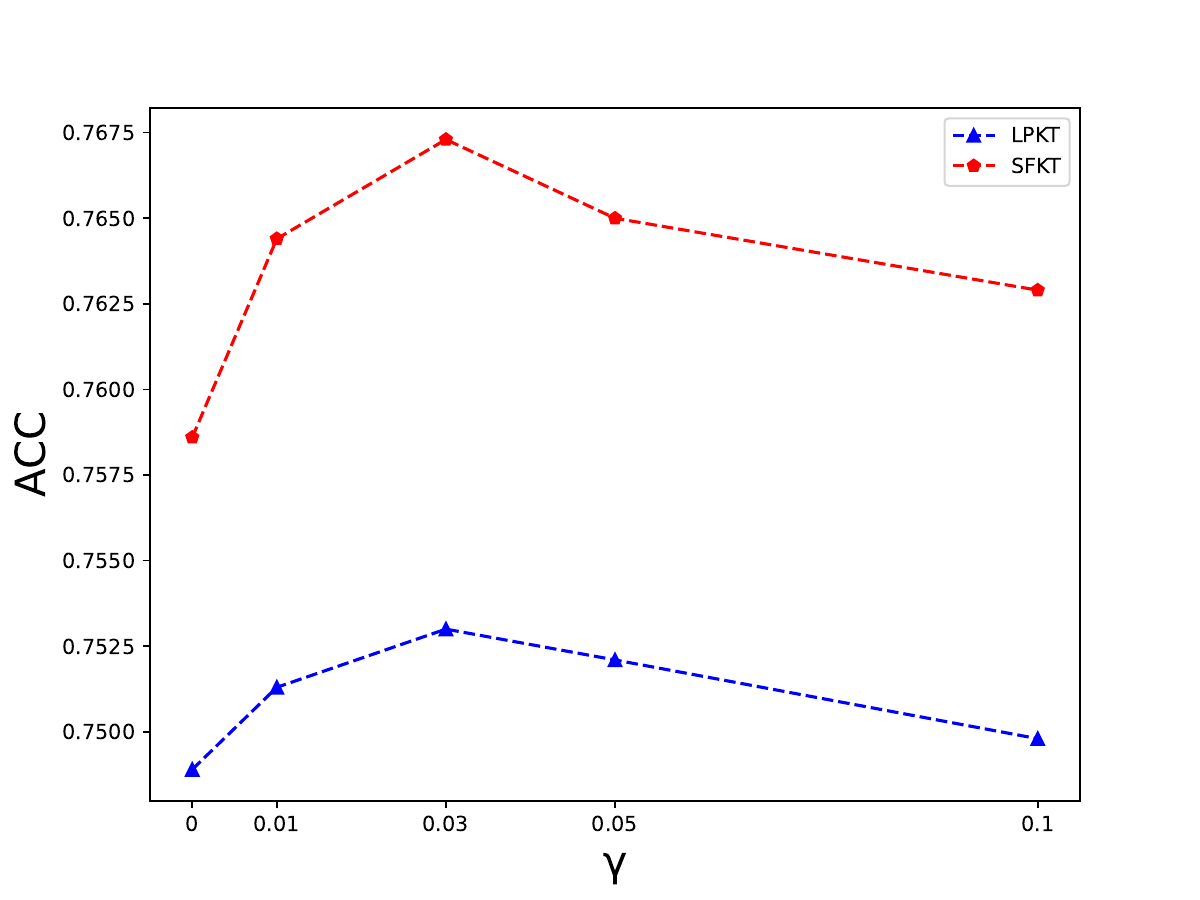}
  \end{minipage}
  \hfill
  \begin{minipage}[b]{0.23\textwidth}
    \centering
    \includegraphics[width=\linewidth]{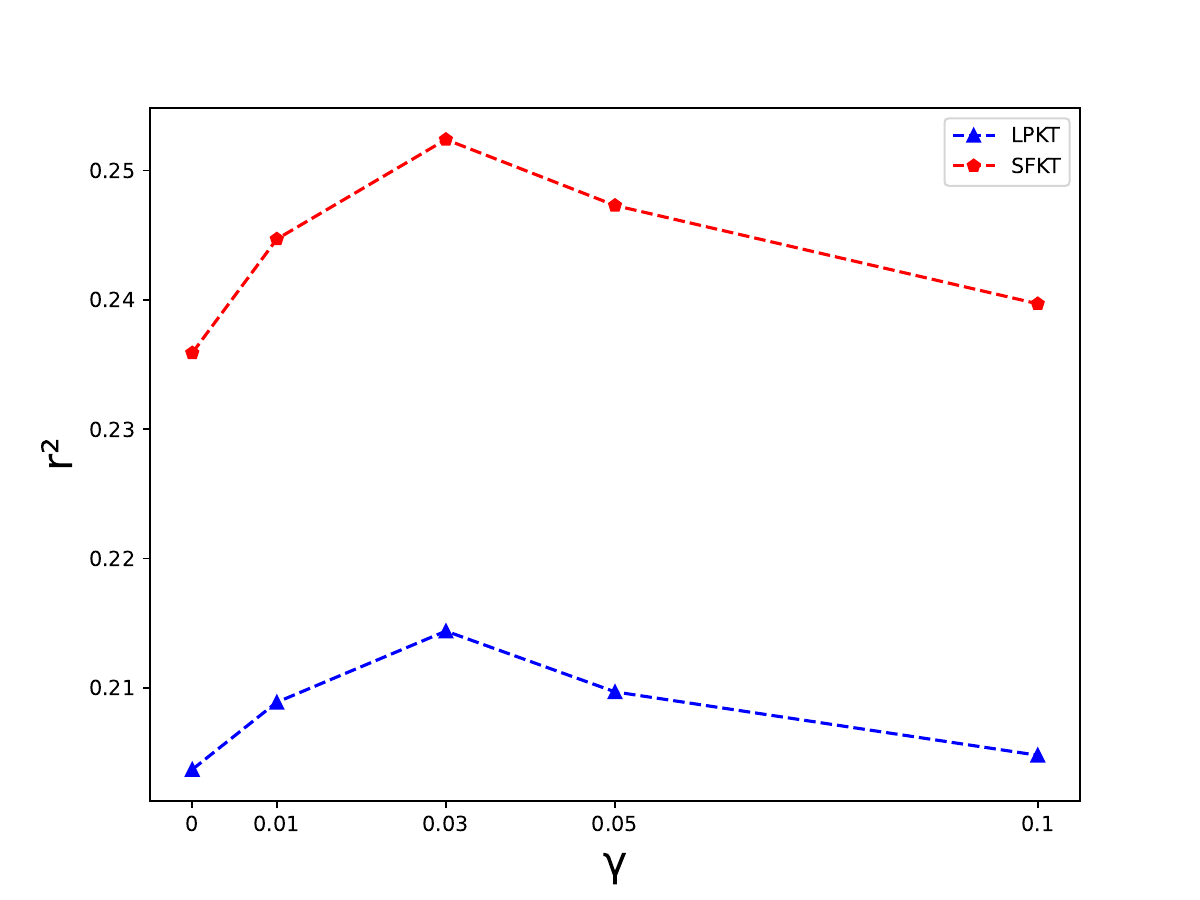}
  \end{minipage}
  \caption{The influence of the small positive value $\bm{\gamma}$ in the enhanced Q-matrix on the performance of LPKT and \methodname on ASSIST2012.}
  \label{fig:param}
\end{figure}
\subsubsection{Parameter Sensitivity}
Additionally, we conducted experiments to assess the impact of $\gamma$ on the LPKT and \methodname enhanced Q matrix. We tried five different $\gamma$ values: $0$, $0.01$, $0.03$, $0.05$ and $0.1$. The experimental results are shown in Fig.~\ref{fig:param}. It can be observed that setting $\gamma$ to a small positive value enhances the performance of LPKT and \methodname, with the maximum gain achieved when $\gamma$ is set to 0.03. When $\gamma$ approaches zero, it becomes challenging for $\gamma$ to bridge the potential correlation between different knowledge concepts. On the other hand, if $\gamma$ increases, more errors may occur in the enhanced Q matrix, which would impair the performance of LPKT and \methodname. From the graph, it is evident that \methodname and LPKT exhibit distinct slopes. In the \methodname model, we first use the Q matrix to match the knowledge concepts present in the current exercise and then utilize the predecessor-successor knowledge concept matrix to identify the prerequisite knowledge concepts of the current one. This step emphasizes that our model relies more on the relationships between knowledge concepts when computing the forgetting process, further validating the effectiveness of the concept-driven forgetting mechanism proposed in our work.

\begin{figure}[ht]
  \centering
  \resizebox{0.35\textwidth}{!}{%
  \begin{minipage}[b]{0.4\textwidth}
    \centering
    \small 
    \begin{tabular}{c|ccccc}
      \hline
      \diagbox[height=2.8em,width=6em]{Metrics}{K} & 0 & 10 & 30 & 50 & 100 \\
      \hline
      AUC &0.8126 & 0.8169 & 0.8193 & $\textbf{0.8206}$ & 0.8182 \\
      ACC &0.7458 & 0.7463 & 0.7482 & $\textbf{0.7495}$ & 0.7478\\
      RMSE &0.4157 & 0.4135 & 0.4119 & $\textbf{0.4075}$ & 0.4130 \\
      r² &0.2760 & 0.2783 & 0.2867 & $\textbf{0.2921}$ & 0.2851 \\
      \hline
    \end{tabular}
  \end{minipage}%
  }
  \hspace{1.95cm} 
  \begin{minipage}[b]{0.25\textwidth}
    \centering
    \includegraphics[width=1.01\linewidth]{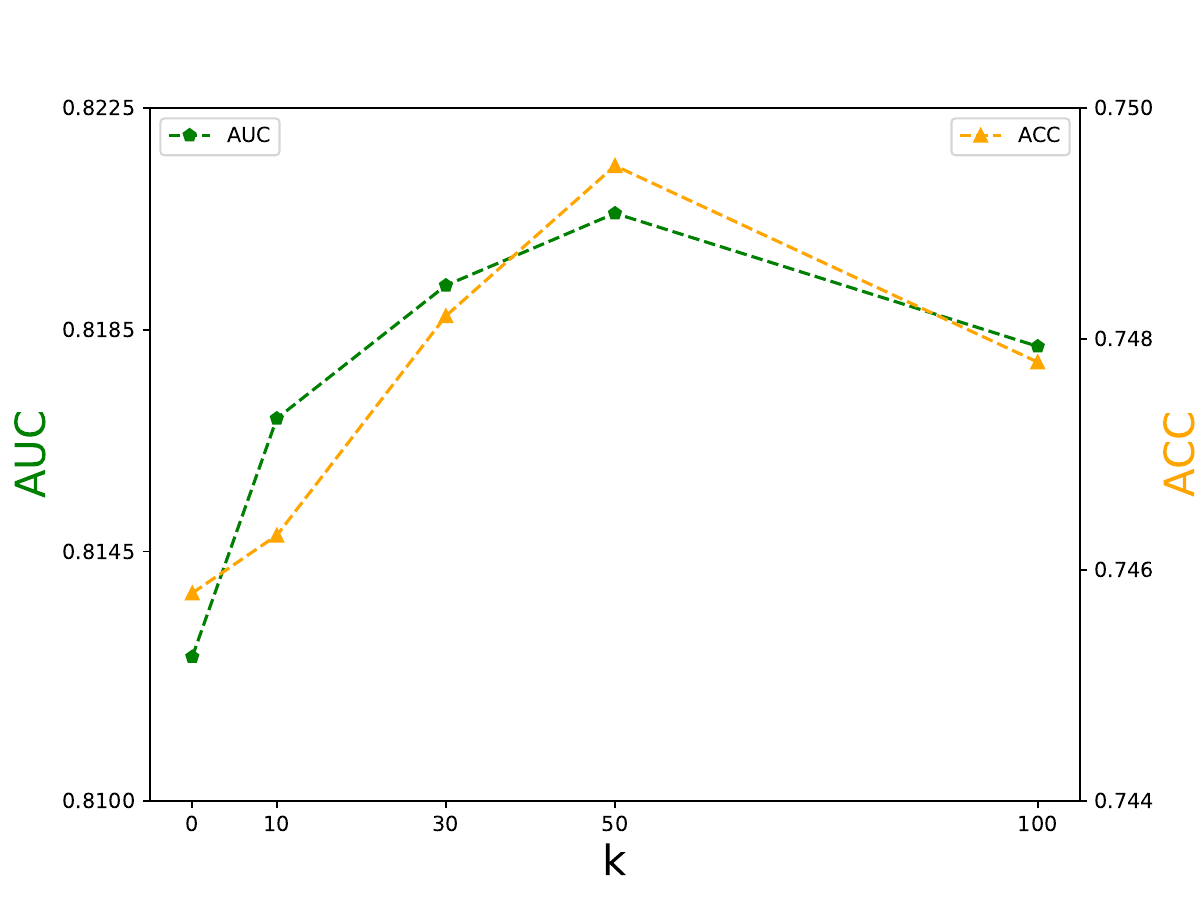}
    \vspace{-5.12em} 
  \end{minipage}%
  \hfill
  \begin{minipage}[b]{0.25\textwidth}
    \centering
    \includegraphics[width=0.93\linewidth]{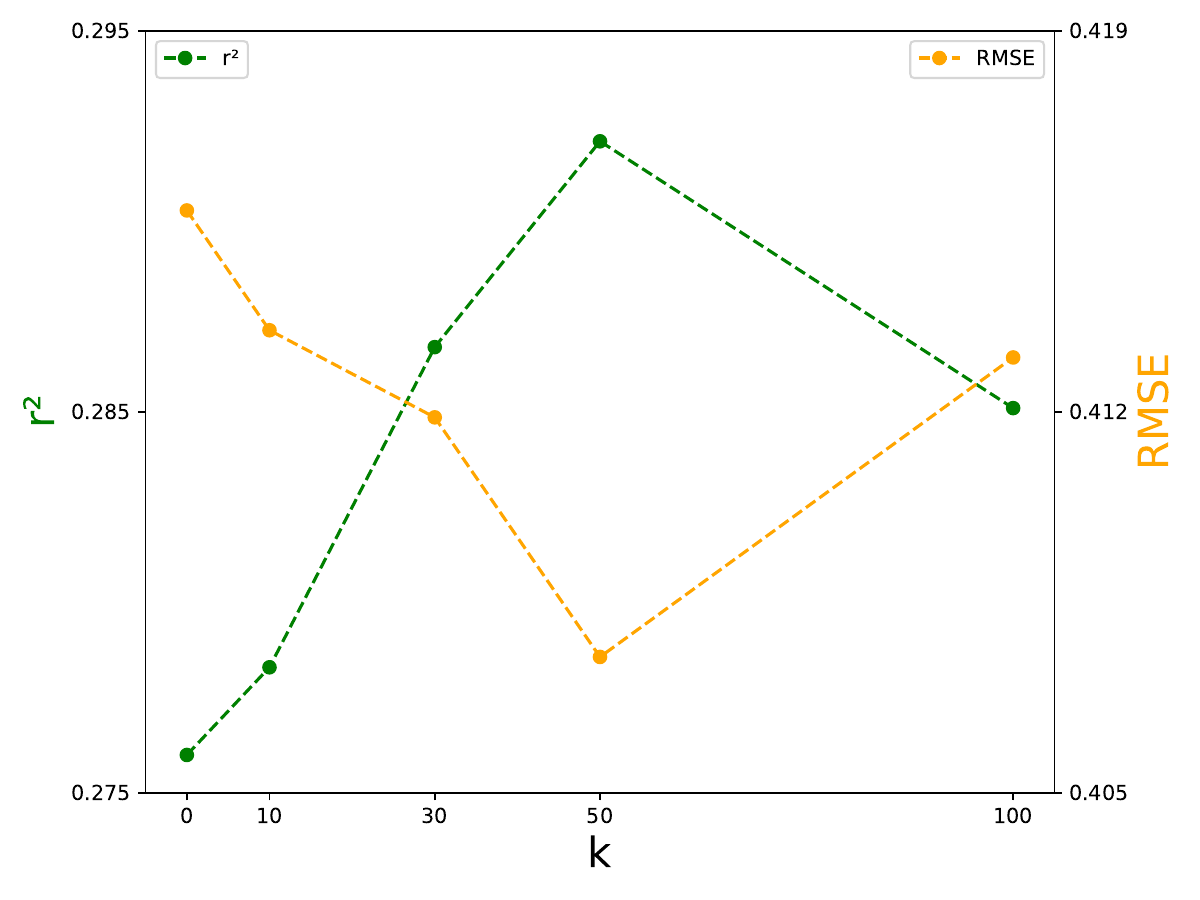}
    \vspace{-3.99em} 
  \end{minipage}
  \vspace{0.5em}
  \caption{Experimental results with different K values on ASSISTchall.}
  \label{fig:diagonal_line_and_pdfs}
\end{figure}

Fig.~\ref{fig:diagonal_line_and_pdfs} illustrates the impact of different values of $k$ on the prediction results when considering the preceding $k$ neighboring knowledge states. Due to the presence of numerous knowledge concepts in practical applications, considering the similarity between all knowledge concepts in the sequence may lead to an increase in computational complexity. Therefore, we experimented with five different values of $k$: 0, 10, 30, 50, and 100. When $k$ is small, it is equivalent to disregarding the memory enhancement process described above. As $k$ increases, the complexity of the relationships between knowledge concepts grows, potentially involving a large number of knowledge concepts in the enhancement process, which could destabilize the model.

\section{Conclusion}
In this paper, we propose a novel concept-driven personalized forgetting knowledge tracking model (\methodname) aimed at addressing personalized learning and forgetting processes in real-world applications. \methodname integrates students' personalized capabilities into both the learning and forgetting processes and models the causal relationships of forgetting processes through the hierarchical relationships between knowledge concepts. Extensive experiments on three public datasets demonstrate the superiority of \methodname over existing methods. Future research directions include further strengthening the \methodname model, considering the multi-concept effects, more comprehensively characterizing the intrinsic relationships between knowledge concepts, and delving deeper into the impact of knowledge concepts on the model.










\end{document}


\ArticleType{Supplementary File}

\title{Title}{Title for citation}

\author[1]{Aaa AUTHOR}{}
\author[1,2]{Bbb AUTHOR}{{bauthor@xxx.com}}
\author[2]{Ccc AUTHOR}{}
\author[3]{Ddd AUTHOR}{}

\AuthorMark{Author A}

\AuthorCitation{Author A, Author B, Author C, et al}


\address[1]{Affiliation, City {\rm 000000}, Country}
\address[2]{Affiliation, City {\rm 000000}, Country}
\address[3]{Affiliation, City {\rm 000000}, Country}

\maketitle


\begin{appendix}

\section{Importance}
Please use this sample as a guide for preparing your letter. Please read all of the following manuscript preparation instructions carefully and in their entirety. The manuscript must be in good scientific American English; this is the author's responsibility. All files will be submitted through our online electronic submission system at \href{https://mc03.manuscriptcentral.com/scis}{HERE}.

\section{More information}
The examples at the bottom of the .tex file can help you when preparing your manuscript. We are appreciate your effort to follow our style~\cite{1,2}.

\end{appendix}
